\documentclass[10pt,twocolumn,letterpaper]{article}

\usepackage[pagenumbers]{cvpr} % To force page numbers, e.g. for an arXiv version

\usepackage{graphicx}
\usepackage{enumitem}
\usepackage{multirow}
\usepackage{colortbl}
\usepackage{booktabs}
\usepackage{amsmath}
\usepackage{amssymb}
\usepackage{xspace}
\usepackage{soul}
\usepackage{comment}
\usepackage{epigraph}
\usepackage{algorithm}
\usepackage{algpseudocode}
\usepackage{tcolorbox}
\usepackage{pifont}
\usepackage{cuted}
\usepackage[accsupp]{axessibility}
\usepackage{caption}
\usepackage{hyperref}
\usepackage{url}

\definecolor{Gray}{gray}{0.90}

\newcommand{\bhdr}[1]{\noindent\textbf{#1}}

\def\modelname{DAWN\xspace}

\def\systwoname{Motion Director\xspace}
\def\sysonename{Action Expert\xspace}

\definecolor{skyblue}{rgb}{0.04,0.40,0.80}
\hypersetup{
    colorlinks,
    linkcolor={red},
    citecolor={skyblue},
    urlcolor={magenta}
}

\definecolor{forestgreen}{rgb}{0.13,0.55,0.13}

\definecolor{cvprblue}{rgb}{0.21,0.49,0.74}
\def\confName{CVPR}
\def\confYear{2026}

\title{Pixel Motion Diffusion is What We Need for Robot Control}

\author{%
E-Ro Nguyen\thanks{Equal contribution.}\quad
Yichi Zhang\footnotemark[1]\quad
Kanchana Ranasinghe\quad 
Xiang Li\quad
Michael S. Ryoo \\
Stony Brook University\\
\texttt{eronguyen@cs.stonybrook.edu} \\
\href{https://eronguyen.github.io/DAWN}{\texttt{eronguyen.github.io/DAWN}
}
}

\begin{document}
\maketitle

% \begin{figure*}[h]
%     \centering
%     \includegraphics[width=0.9\linewidth]{Images/overview_2.pdf}
%     \caption{Overview of \modelname with two major diffusion modules. First, observations are encoded into conditional embeddings; Based on that, a latent diffusion Motion Director generates a pixel motion representation, which the diffusion policy Action Expert uses to create robot actions.}
%     \label{fig:arch}
% \end{figure*}

\begin{strip}
\begin{minipage}{\textwidth}\centering
\vspace{-10pt}
\includegraphics[width=0.85\textwidth]{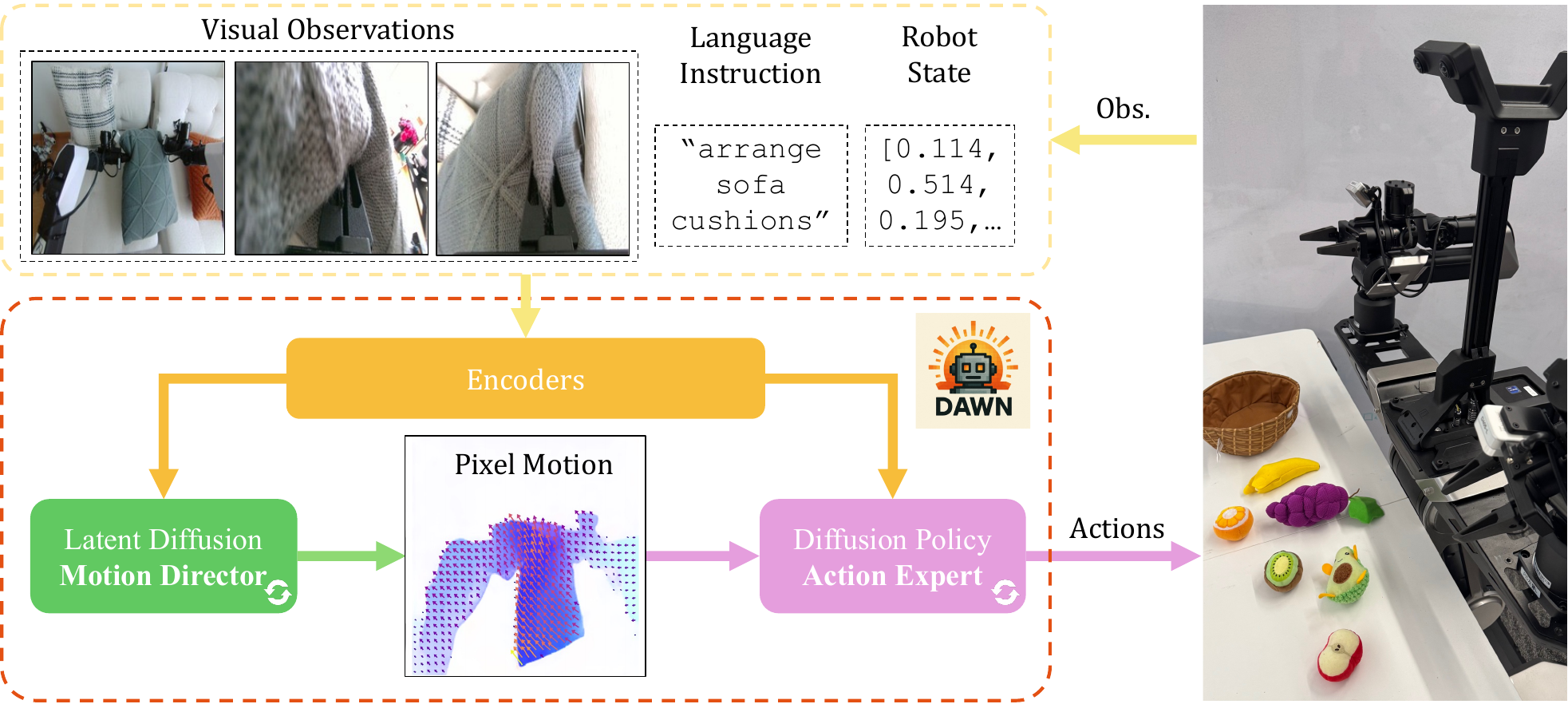}
\captionof{figure}{Overview of \modelname with two major diffusion modules. Given visual observations, robot state, and a language instruction, a latent diffusion Motion Director predicts a dense pixel motion representation that describes the desired scene dynamics, which the diffusion policy Action Expert uses to generate executable robot actions. Explicit pixel motion provides a structured, interpretable interface between perception and control.
% First, observations are encoded into conditional embeddings; Based on that, a latent diffusion Motion Director generates a pixel motion representation, which the diffusion policy Action Expert uses to create robot actions.
}
  \label{fig:teaser}
\end{minipage}
\end{strip}

\begin{abstract}
We present \modelname (Diffusion is All We Need for robot control), a unified diffusion-based framework for language-conditioned robotic manipulation that bridges high-level motion intent and low-level robot action via structured pixel motion representation. In \modelname, both the high-level and low-level controllers are modeled as diffusion processes, yielding a fully trainable, end-to-end system with interpretable intermediate motion abstractions. 
\modelname achieves state-of-the-art results on the challenging CALVIN benchmark, demonstrating strong multi-task performance, and further validates its effectiveness on MetaWorld. Despite the substantial domain gap between simulation and reality and limited real-world data, we demonstrate reliable real-world transfer with only minimal finetuning, illustrating the practical viability of diffusion-based motion abstractions for robotic control. Our results show the effectiveness of combining diffusion modeling with motion-centric representations as a strong baseline for scalable and robust robot learning. 

% Visualization page:  \href{https://anonymous.4open.science/w/DAWN}{\texttt{anonymous.4open.science/w/DAWN}}.
\end{abstract}

\section{Introduction}
\label{sec:intro}
Recent vision-language-action (VLA) models~\citep{Black20240AV,Intelligence2025pi05,kim2024openvla,Zhang2025DreamVLAAV,black2024pi_0}
have achieved strong generalization by leveraging large-scale web and robot datasets. 
Some works~\citep{Li2024LLaRASR,niu2024llarva,yang2025magma,zheng2024tracevla,Zawalski2024RoboticCV} suggest that it is beneficial to include explicit spatial reasoning and motion awareness into the action generation process.
% However, these models typically operate on latent tokens, leaving them without an explicit representation of how objects or the robot’s end-effector should move to accomplish an instruction. % The absence of structured motion reasoning limits their reliability in environments requiring precise physical interaction.
% Our method provides an alternative way to present the fine-grained robot motion
Several approaches~\citep{Yuan2024GeneralFA,Gao2024FLIPFG,Xu2024FlowAT,Bharadhwaj2024Track2ActPP,Bharadhwaj2024Gen2ActHV} introduce multi-stage pixel or point tracking to form interpretable intermediate motion cues. % While promising, these methods often rely on object detectors, bounding boxes, or keypoints~\citep{Xu2024FlowAT,singh2025afford2act}, which make them limited under occlusions, or scenes without well-defined object boundaries.

More recently, predictive representations based on video diffusion and GPT-style forecasting models ~\citep{Hu2024VideoPP,wu2023unleashing} are utilized to predict future observations directly.  Gen2Act~\citep{Bharadhwaj2024Gen2ActHV} extracts motion trajectories by tracking pixels within their generated videos. 
The motion trajectories offer fine-grained guidance on intended movements, which has been empirically shown to improve robotic control. This insight motivates a simplified approach in which we directly predict dense pixel motions instead of RGB video frames, thereby reducing complexity and making the underlying learning problem easier for the network.

% The motion trajectories provide relatively fine-grained / details/ dense motion intention guidance, which is empirically beneficial for the robot control. 
% [motivation], simpler, making it easier to learn 

% However, when motion is derived implicitly from generated videos, the predictions inevitably inherit temporal inconsistency and misalignment present in those frames. 
% This results in unstable motion cues that are difficult to use for precise, closed-loop robotic control. 
% (P2-3 tone is too strong)

Therefore, we propose \textbf{D}iffusion is \textbf{A}ll \textbf{W}e \textbf{N}eed for robot control (\modelname), a two-stage diffusion-based visuomotor framework that predicts \emph{dense pixel motion explicitly} rather than deriving it indirectly from generated videos, detectors, or keypoints. Both the high-level and low-level controllers are implemented as diffusion models and are connected through structured pixel motion, forming an interpretable and modular control pipeline (see~\Cref{fig:teaser}). Our Motion Director uses a latent diffusion model to generate dense pixel motion fields conditioned on current observations and language instruction. These pixel motions serve as an intermediate representation of desired scene dynamics to accomplish the language instruction. 
The Action Expert, a diffusion-based policy model, then translates these motions to executable robot actions, forming a coherent control framework. Our approach bridges the strengths of hierarchical motion decomposition and end-to-end visuomotor agents, while maintaining interpretability and modularity. 

We evaluate our method on two challenging simulation benchmarks—CALVIN~\citep{mees2022calvin} and MetaWorld~\citep{yu2019meta}, as well as across real-world environments with only very limited in-domain training data.

Our results demonstrate that, despite using limited data and substantially smaller model capacity, our method can match or even surpass state-of-the-art VLA models by leveraging explicit structured pixel motion and the strengths of diverse pretrained models, highlighting its high data efficiency.

\noindent
Our key contributions are as follows:
\begin{enumerate}[leftmargin=1.2em,noitemsep,topsep=-0.0em,itemsep=-0.0ex,partopsep=0ex,parsep=1ex]
\item We propose \modelname, a two-stage
diffusion-based framework that generates structured intermediate pixel motion as an efficient language-conditioned visuomotor policy.
\item Despite relying on limited data and a substantially smaller model capacity, we achieve competitive or even state-of-the-art performance on CALVIN, MetaWorld, and real-world benchmarks.
\item Our approach is explicitly designed to leverage pretrained vision and language models, enabling highly data-efficient transfer across domains, while providing interpretability and modularity. 
\end{enumerate}

\begin{figure}
    \centering
    \includegraphics[width=1\columnwidth]{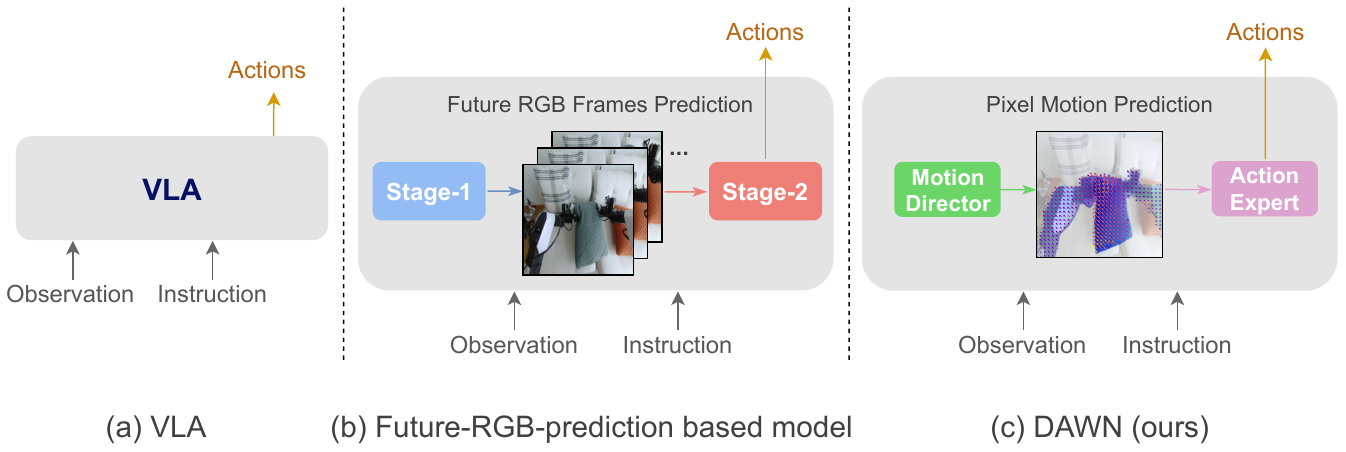}
    \caption{\textbf{Comparison of Action-Prediction Frameworks.} a) VLA directly maps observations and language instructions to action outputs. b) Future-RGB-frame-prediction first generate future visual observations and subsequently condition the action policy on these predicted frames. c) Our proposed \modelname predicts pixel-motion representations via the \systwoname and converts them into actions using the \sysonename, enabling a more informative and structured intermediate representation for action prediction.}
    \label{fig:model_compare}
    \vspace{-5mm}
\end{figure}        %         

\begin{figure*}[t]
    \centering
    \begin{minipage}[t]{0.48\linewidth}
        \centering
        \includegraphics[width=0.9\linewidth]{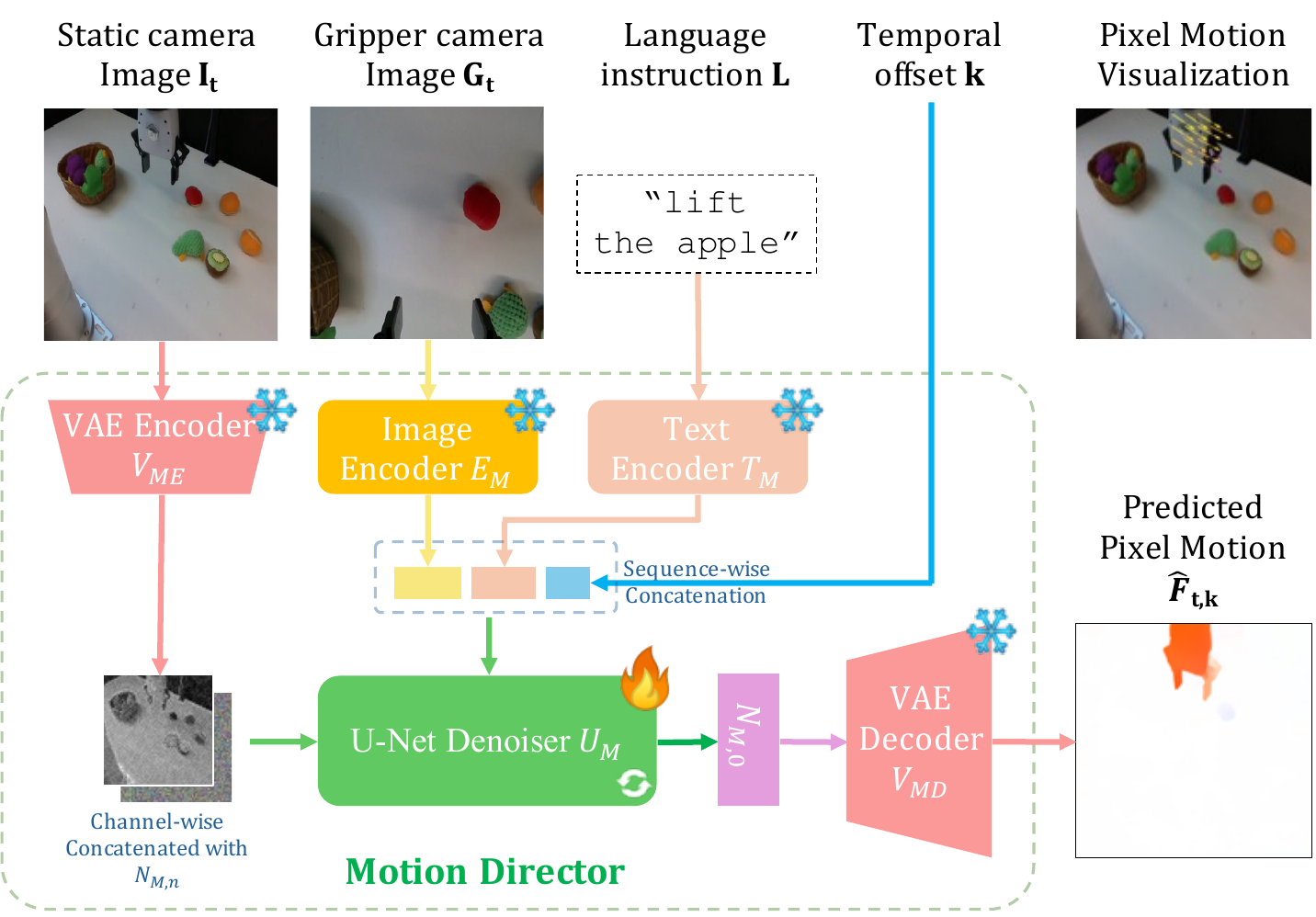}
        \caption{Architecture of \systwoname. The model encodes the static camera view and denoises it with a U-Net, conditioned on the gripper view, language instruction with a temporal offset. The output is decoded into predicted pixel motions, providing interpretable motion representations.}
        \label{fig:system2_arch}
    \end{minipage}\hfill
    \begin{minipage}[t]{0.48\linewidth}
        \centering
            \includegraphics[width=0.87\columnwidth]{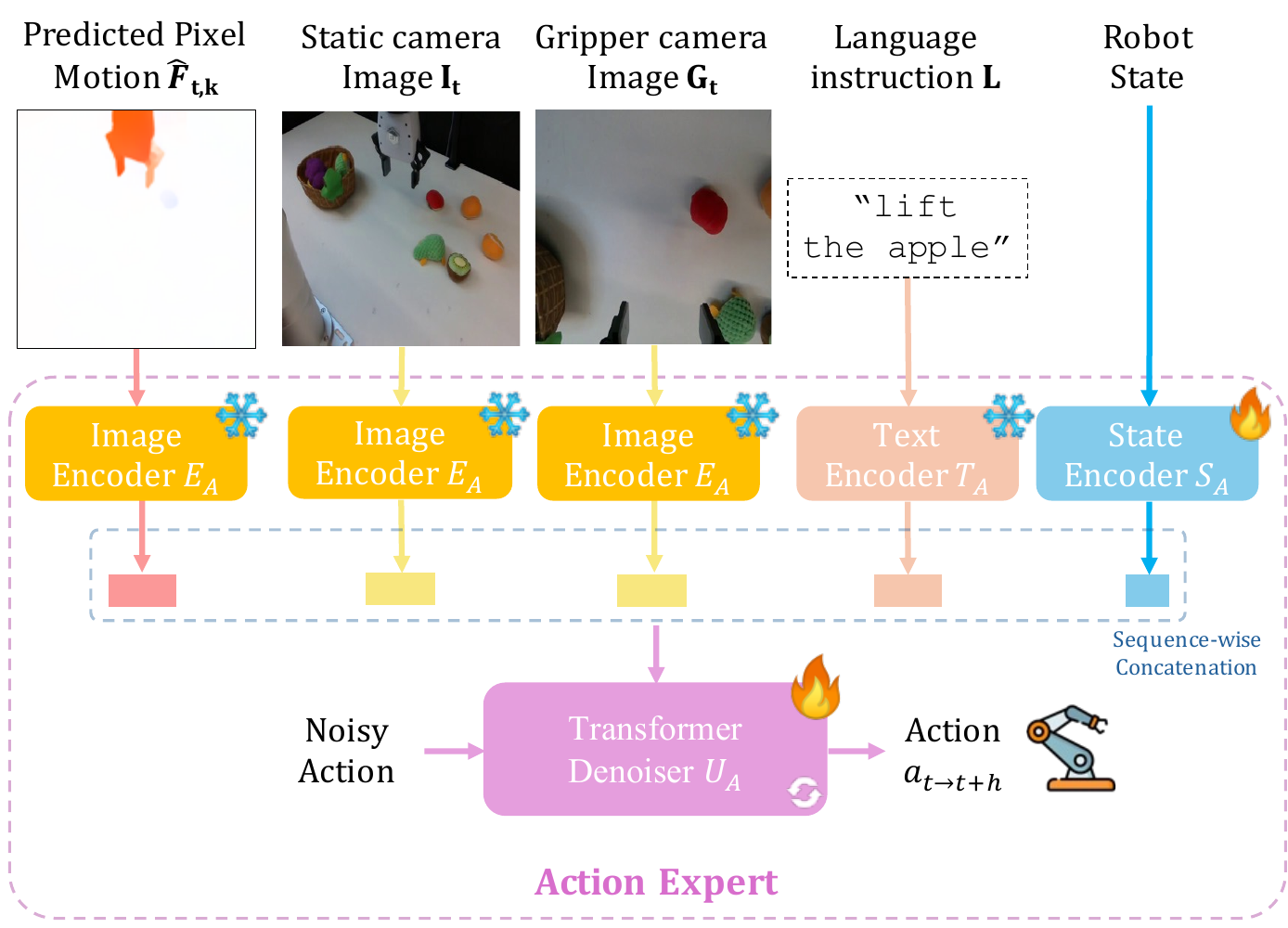}
            \caption{Architecture of \sysonename. The model encodes predicted pixel motion, visual observations, language instruction, and robot state into multimodal features. These inputs condition the denoising process, which iteratively refines noisy actions into executable robot trajectories.}
            \label{fig:system1_arch}

    \end{minipage}
    \vspace{-2em}
\end{figure*}

\section{Related Work}
\label{sec:related}
\bhdr{Sparse Pixel Trajectories for Robot Control:}
Several prior works explore pixel trajectories as motion representations~\citep{Bharadhwaj2024Track2ActPP,Bharadhwaj2024Gen2ActHV,Hu2024VideoPP,Ran25LTM}. These methods capture the displacement of pixels between consecutive frames, providing a local description of motion that is universal and often embodiment-agnostic.
For instance, General Flow~\citep{Yuan2024GeneralFA} proposes a language-conditioned sparse 3D flow prediction model trained on large-scale human videos and treats sparse 3D flow as a foundational affordance, providing a scalable, universal language for describing manipulation. FLIP~\citep{Gao2024FLIPFG} utilizes a flow-centric generative planning model to synthesize long-horizon plans from language-annotated videos, guiding low-level policies. Im2Flow2Act~\citep{Xu2024FlowAT} and Track2Act~\citep{Bharadhwaj2024Track2ActPP} both use point or object flow as a cross-domain interface, bridging the gap between human videos, simulated data, and real-world robot execution to achieve zero-shot manipulation. 
% Finally, Gen2Act~\citep{Bharadhwaj2024Gen2ActHV} takes a generative approach, first imagining a video of future motion in image pixel space and then conditioning a robot policy on the generated video to enable generalizable manipulation. 
AVDC~\citep{Ko2023LearningTA} and Gen2Act~\citep{Bharadhwaj2024Gen2ActHV} adopt generative approaches that synthesize future observations and derive motion signals (such as optical flow or point trajectories) to guide robot policies. 

Pixel trajectories are also found useful when presented in different modalities like image or text.
GENIMA~\citep{Shridhar2024GenerativeIA} fine-tunes a diffusion model to inpaint markers on visual observations, which could be decoded into robot actions. 
LLaRA~\citep{Li2024LLaRASR} presents the robot action in text-based image pixel coordinates and formats the robot policy into a conversation style to benefit from a pretrained large VLM. This enables an efficient transfer from a general VLM into VLA. Similarly, RoboPoint~\citep{yuan2024robopoint}, LLARVA~\citep{niu2024llarva}, TraceVLA~\citep{zheng2024tracevla}, Magma~\citep{yang2025magma} and HAMSTER~\citep{li2025hamster} all take advantage of different kinds of textual pixel trajectories.

% rely on sparse motion representations and often depend on additional components such as object detectors, point selection mechanisms, or tracking pipelines to obtain task-relevant motion signals

Unlike these methods that depend on sparse pixel trajectories and often depend on additional components such as object detectors, point selection mechanisms, our framework predicts high-resolution dense pixel motion during action generation using a latent diffusion model pretrained on general images.

% Despite their effectiveness, many of these approaches rely on sparse motion representations or additional components such as object detection, point tracking, or keypoint selection. In contrast, our framework directly predicts \textit{dense pixel motion} using a latent diffusion model conditioned on current observations and language instructions. By formulating motion prediction as a latent image-generation problem, our method produces high-resolution motion fields that capture fine-grained scene dynamics and provide richer guidance for action generation.
% Compared with prior motion-based approaches such as LangToMo~\citep{Ran25LTM}, which predicts low-resolution motion from a single static view, our method generates higher-quality dense motion representations and integrates them into a diffusion-based action policy. As shown in our experiments, this leads to more accurate motion prediction and improved manipulation performance across benchmarks.

\bhdr{Future RGB Frame Prediction for Robot Control:}
Recent work has increasingly treated powerful generative video and image models as world models or planners for robot manipulation.
SuSIE~\citep{Black2023ZeroShotRM} finetunes an image-editing diffusion model to generate subgoal images from current observations and commands, then realizes it, enabling strong zero-shot generalization.
UniPi~\citep{du2023unipi} uses a video-diffusion model to plan future frames from language goals and then iinfers actions via inverse dynamics. 
% FlowDreamer~\citep{guo2025flowdreamer} predicts 3D scene flow and uses a diffusion model to render future RGB-D observations for planning. 
VidMan~\citep{wen2024vidman} pretrains a video-diffusion model on large-scale human data and then adapts it into an inverse-dynamics action model.
VPP~\citep{Hu2024VideoPP} employs a video diffusion model to extract predictive feature embeddings, which subsequently condition a downstream action policy. 

However, these approaches operate purely in RGB space without an explicit motion-centric representation, whereas our method directly predicts dense pixel motion (see~\Cref{fig:model_compare}). 

\bhdr{Vision-Language-Action Models:}
Vision-language-action models have emerged as a powerful paradigm for language-conditioned robot control~\citep{rt1,rt2,bahl2022human,padalkar2023open,reed2022gato,wu2023unleashing,driess2023palm,kim2024openvla,zheng2024tracevla,Zawalski2024RoboticCV,Sudhakar2024ControllingTW,Jeong2025ObjectCentricWM,yang2025magma}. Leveraging large-scale training with web-scale vision-language data, these models increasingly focus on improving generalization and data efficiency. 

RT-2~\citep{rt2} jointly finetuned a vision–language model on robot trajectory data and large-scale web image/text tasks. The key idea was to treat robot actions as discrete tokens in the same vocabulary as language tokens, thus enabling a unified modelling framework.
OpenVLA~\citep{kim2024openvla} continued the discrete-token formulation of actions and emphasized parameter-efficient fine-tuning for new robot setups.
$\pi$ series~\citep{black2024pi_0, Intelligence2025pi05} allows higher fidelity, higher‐frequency control with the help of the flow-matching based action expert.

% Spatial reasoning has recently become central to VLA design. 
% Spatial-VLA~\citep{qu2025spatialvla} introduces spatial tokenization and 3D grounding layers so that the model explicitly predicts where to act by operating over 3D feature volumes. This line of work highlights that visual–language representations alone inadequately encode actionable spatial cues, motivating architectures that integrate geometry or 3D fields.
% GR00T N1~\citep{bjorck2025gr00t} incorporates structured 3D scene encodings and hierarchical skill decoders. 

% Inspired partly by game-engine simulation and 3D world models, GR00T emphasizes spatial grounding for long-horizon tasks and multi-dozen‐step instructions.

% DVD~\citep{chen2021learning} uses diverse ``in-the-wild" human videos to teach reward functions and enables zero-shot transfer to new environments. 
% GR-1~\citep{wu2023unleashing} is a GPT-style transformer policy that benefits from large-scale video pre-training.   
% 3D-VLA~\citep{zhen20243d} proposes a world model that integrates 3D perception and reasoning to enhance planning capabilities. These advances have been supported by initiatives like Octo~\citep{octo_2023}, an open-source generalist policy trained on the vast Open X-Embodiment dataset~\citep{o2024open}, paving the way for more reproducible and widely usable models. 

\section{Method}
\label{sec:dawn}

\subsection{Preliminaries: Diffusion Models}
Diffusion models~\citep{sohl2015deep, ho2020denoising, ddim} are powerful generative models that synthesize data by iteratively denoising noise-corrupted inputs to approximate the target data distribution. The process involves a forward step that gradually perturbs real data with noise through a Markov chain, and a reverse step in which a neural network parameterizes Gaussian transitions to progressively remove noise, eventually generating realistic samples from a simple distribution such as a standard Gaussian. These models have shown remarkable success in image generation and broader data generation tasks. 
To improve scalability and efficiency for image generation, latent diffusion models~\citep{rombach2022high} operate in a compressed latent space rather than raw pixel space, significantly reducing computational demands while preserving fidelity. 

Diffusion-based approaches have also been adapted for robot learning, where policy learning can be framed as a sequence generation problem. Diffusion Policy~\citep{Chi2023DiffusionPV}, in particular, addresses the challenge of visuomotor control by generating action sequences conditioned on both visual and low-dimensional states. %

\subsection{Problem Formulation and \modelname Overview}
We study the topic of language-instructed visuomotor control, where the goal is to build a policy that takes both visual observations and natural language instructions from the environment to generate robot actions for control, using behavior cloning (i.e., supervised learning).

Our approach, \modelname, combines the strengths of two complementary diffusion models: a latent image diffusion model for pixel-level motion generation, referred to as \textit{\systwoname}, and a diffusion transformer for fine-grained action sequence generation, referred to as \textit{\sysonename}. These two models interact through explicit pixel-motion representations. 
At a higher level, \systwoname conditions on multi-view images and the language instruction to iteratively generate task-aligned pixel motions, grounded to one of the input views as illustrated at~\Cref{fig:system2_arch}. At the lower level, \sysonename takes the generated pixel motion along with additional inputs to produce the final robot action sequence as illustrated at~\Cref{fig:system1_arch}. 
We highlight how our pixel-motions are grounded to an input view, endowing our intermediate representations with interpretability. 

\subsection{Motion Director}

Consider two videos $\mathbf{I}, \mathbf{G} \in \mathbb{R}^{T \times H \times W \times C}$ capturing the same robot demonstration from different camera views, each consisting of $T$ frames of height $H$, width $W$, and $C$ channels, along with the corresponding language instruction $\mathbf{L}$. For example, $\mathbf{I}$ could be the video from a static third-person view, and $\mathbf{G}$ could be captured from the camera above the gripper.
Let $\mathbf{I}_t, \mathbf{G}_t$ denote the $t$-th frame from the corresponding views. We define the pixel motion from $\mathbf{I}_t$ to $\mathbf{I}_{t+k}$ as $\mathbf{F}_{t,k} = [u, v]$, where $u, v \in \mathbb{R}^{H \times W}$ represent amount of movement of each pixel between $\mathbf{I}_t$ and $\mathbf{I}_{t+k}$ in the horizontal and vertical directions, respectively. To take advantage of pretrained models, we further encode this motion into a three-channel image $\mathbf{F'}_{t,k} = [u, v, (u+v)/2]$.

The goal of Motion Director is to estimate $\mathbf{F'}_{t,k}$ using only current visual input $\mathbf{I}_t$, $\mathbf{G}_t$ and instruction $\mathbf{L}$. Our \systwoname builds on a pretrained latent diffusion model for RGB image generation, comprising a U-Net denoiser~$U_M$, a text encoder~$T_M$, and pretrained VAE encoder–decoder pair ($V_{ME}$, $V_{MD}$). We also incorporate a vision encoder~$E_M$ to extract embeddings from alternative camera views.

At the inference time, we first draw a Gaussian noise $\mathbf{N}_{M,n}$ and concatenate it with the latent encoding of the current frame $V_{ME}(\mathbf{I}_t)$, forming a noisy latent representation $\mathbf{O}_{M,n}=[N_{M,n}, V_{ME}(\mathbf{I}_t)]$, where $n$ is the total number of denoising steps we plan to execute. Note that the current frame latent encoding $V_{ME}(\mathbf{I}_t)$ does not undergo any form of corruption, as this is a conditioning signal. 
The U-Net~$U_M$ then denoises $\mathbf{O}_{M,n}$ and outputs a less noisy latent $N_{M,n-1}$ under the conditioning of the language embedding $T_M(\mathbf{L})$, visual embedding of the alternative view $E_M(\mathbf{G}_t)$, and a temporal offset $k$. 
All conditioning tokens are concatenated and injected into the U-Net’s cross-attention layers at each denoising step. The denoised latent tensor will be concatenated again with the VAE encoded visual inputs to form the input for the next denoising step $\mathbf{O}_{M,n-1}=[N_{M,n-1}, V_{ME}(\mathbf{I}_t)]$. For an arbitrary denoising step $i$, the process can be presented as below, where $t_i$ is the denoising timestamp and $\left[... \right]$ stands for concatenation.
\begin{align}
    \mathbf{O}_{M,i} &=\left[N_{M,n}, V_{ME}(\mathbf{I}_t)\right] \\
    \mathbf{C}_{M} &= \left[E_M(\mathbf{G}_t), T_M(\mathbf{L}), k \right]\\
    \mathbf{N}_{M,i-1} &=U_M(\mathbf{O}_{M,i}, \mathbf{C}_{M}, t_i)
    \label{eq:denoise_motion}
    % \vspace{-10mm}
\end{align}
After $n$ iterations, the denoised latent tensor $\mathbf{N}_{M,0}$ is decoded by $V_{MD}$ into a three-channel image, which ideally matches the ground-truth motion $\mathbf{F'}_{t,k}$.

During training, we update only the U-Net denoiser~$U_M$, while keeping all other modules frozen. The ground-truth pixel motion corresponding to frame $\mathbf{I}_t$ is obtained using the optical flow model RAFT~\citep{teed2020raft} since we have access to future frames during training (i.e., using frames $\mathbf{I}_t$ and $\mathbf{I}_{t+k}$ as input to RAFT), and subsequently projected into latent space through the VAE encoder~$V_{ME}$.

\subsection{Action Expert}

Our \sysonename is responsible for translating pixel motions into low-level robot actions, conditioned on visual observations, robot states, and language instructions. 
To achieve this, motivated by prior diffusion based policies~\citep{Chi2023DiffusionPV}, we construct a transformer based Enhanced Diffusion Policy, 
which generates action sequences by progressively denoising noisy action representations under multimodal conditions. This design enables the policy to capture complex dependencies across modalities while producing coherent actions temporally.

The architecture consists of four key components: (1) a shared visual encoder $V_A$ that encodes both the pixel motion output from \systwoname and the current visual observations, (2) a text encoder $T_A$ that embeds the language instruction, (3) a state encoder $S_A$ that processes low-dimensional robot states through a two-layer MLP, and (4) a denoising transformer $U_A$ that generates action sequences. We initialize $U_A$ and $S_A$ from scratch to allow adaptation to the target task, while keeping the pretrained $V_A$ and $T_A$ frozen to benefit from strong pretrained visual and language representations.

During inference, the pixel motion predicted by \systwoname, together with the visual inputs, language instruction, and robot states, are each processed by their corresponding encoder and projected into token embeddings. These context tokens are concatenated to form the conditioning sequence, which is injected into all transformer blocks of the denoising transformer $U_A$ via cross-attention, following the same mechanism as in \systwoname. Action generation begins from a noisy action chunk with length $h$ sampled from a Gaussian prior, which is iteratively denoised by $U_A$ into a coherent sequence of executable robot actions.

\subsection{\modelname Training and Inference}

In summary, both \systwoname and \sysonename are trained with a mean squared error noise estimation loss. While \systwoname operates in the latent image space to predict pixel motions, \sysonename focuses on predicting action chunks in the robot’s action space.

At inference, observations are first encoded, after which \systwoname produces a single pixel-motion field that conditions \sysonename to generate a sequence of robot actions.  
Once these actions are executed, the system repeats the process with the updated observations, thereby forming a closed-loop control pipeline.

One advantage of this design is that the two diffusion models can be trained in parallel using the optical flow between two images as the ground-truth pixel motion. Two modules could be upgraded independently, enabling flexible integration of future advances in vision or control. After that, \sysonename could optionally be further fine-tuned on the actual pixel motions generated by \systwoname for a better performance.

To the best of our knowledge, this is the first work to adapt a pretrained \emph{latent} diffusion model for dense pixel \emph{motion} generation and use the pixel motion to guide a diffusion policy for visuomotor control under fully learnable settings.

\begin{table*}[t]
    \centering
    \begin{minipage}[t]{0.48\linewidth}
        \centering
        
% \begin{table*}[t]
\centering
\small
\def\arraystretch{1.1}  %
\setlength\tabcolsep{0.9em}  %
\caption{
\textbf{CALVIN Evaluation (no external robotic data):} 
Results reported for zero-shot long-horizon evaluation on the Calvin ABC$\rightarrow$D benchmark. All methods are trained only on the CALVIN dataset without any external data. 
}

\scalebox{0.75}{%
\begin{tabular}{ccccccc}
\toprule
\multirow{2}{*}{\textbf{Method}} & \multicolumn{5}{c}{\textbf{$i^{th}$ Task Success Rate}} & \multirow{2}{*}{\textbf{Avg. Len $\uparrow$}} \\ \cline{2-6} 
 & \textbf{1} & \textbf{2} & \textbf{3} & \textbf{4} & \textbf{5} & \\ \midrule
Diffusion Policy~\citep{Chi2023DiffusionPV} & 0.40 & 0.12 & 0.03 & 0.01 & 0.00 & 0.56 \\
Robo-Flamingo~\citep{li2023vision} & 0.82 & 0.62 & 0.47 & 0.33 & 0.24 & 2.47 \\
Moto~\citep{chen2024moto} & 0.90 & 0.73 & 0.60 & 0.48 & 0.39 & 3.10 \\
MoDE~\citep{reuss2025efficient} & 0.92 & 0.79 & 0.67 & 0.56 & 0.45 & 3.39 \\
RoboUniview~\citep{yang2025magma} & 0.94 & 0.84 & 0.73 & 0.62 & 0.51 & 3.65 \\
Seer~\citep{Tian2024PredictiveID} & 0.93 & 0.82 & 0.72 & 0.63 & 0.53 & 3.64 \\
Seer-Large~\citep{Tian2024PredictiveID} & 0.93 & 0.85 & 0.76 & 0.69 & 0.60 & 3.83 \\
VPP~\citep{Hu2024VideoPP} & 0.96 & 0.88 & 0.78 & 0.71 & 0.60 & 3.93 \\ 
\midrule
Enhanced DP (ours) & 0.82 & 0.67 & 0.53 & 0.41 & 0.35 & 2.78 \\
\rowcolor{Gray} 
\modelname (ours) & 0.98 & 0.91 & 0.79 & 0.71 & 0.61 & 4.00 \\  

\bottomrule
\end{tabular}
}
% \end{table*}

        \label{tbl:calvin_only}
    \end{minipage}\hfill
    \begin{minipage}[t]{0.48\linewidth}
        \centering
            
% \begin{table*}[t]
\centering
\small
\def\arraystretch{1.1}  %
\setlength\tabcolsep{0.9em}  %
\caption{
\textbf{CALVIN Evaluation with external robotic data:}
Zero-shot long-horizon evaluation on the Calvin ABC$\rightarrow$D benchmark where agent is asked to complete five chained tasks sequentially based on instructions. 
}

\scalebox{0.65}{%
\begin{tabular}{cccccccc}
\toprule
\multirow{2}{*}{\textbf{Method}}& 
\multirow{2}{*}{\textbf{Additional Data}} 
& \multicolumn{5}{c}{\textbf{$i^{th}$ Task Success Rate}} 
& \multirow{2}{*}{\textbf{Avg. Len $\uparrow$}}
\\ \cline{3-7} 
& & \textbf{1} & \textbf{2} & \textbf{3} & \textbf{4} & \textbf{5} & \\ \midrule
GR-1~\citep{wu2023unleashing} & Ego4D & 0.85 & 0.71 & 0.60 & 0.50 & 0.40 & 3.06 \\ 

Vidman~\citep{wen2024vidman} & OpenX subsets & 0.92 & 0.76 & 0.68 & 0.59 & 0.47 & 3.42 \\ 
LTM~\citep{Ran25LTM} & OpenX subsets & 0.97 & 0.82 & 0.73 & 0.67 & 0.61 & 3.81 \\
Seer~\citep{Tian2024PredictiveID} & DROID & 0.94 & 0.87 & 0.80 & 0.72 & 0.64 & 3.98 \\
MoDE~\citep{reuss2025efficient} & Multiple sources & 0.96 & 0.89 & 0.81 & 0.72 & 0.65 & 4.01 \\
Seer-Large~\citep{Tian2024PredictiveID} & DROID & 0.96 & 0.92 & 0.86 & 0.80 & 0.74 & 4.28 \\
VPP~\citep{Hu2024VideoPP} & Multiple sources & 0.97 & 0.91 & 0.87 & 0.82 & 0.77 & 4.33 \\ 
DreamVLA~\citep{Zhang2025DreamVLAAV} & DROID & 0.98 & 0.95 & 0.90 & 0.83 & 0.78 & 4.44 \\ \rowcolor{Gray}
\modelname (ours) & DROID & 0.98 & 0.92 & 0.81 & 0.75 & 0.64 & 4.10 \\ \bottomrule
\end{tabular}
}

% \end{table*}
            
            \label{tbl:calvin_plus}
    \end{minipage}
    % \vspace{-2em}
\end{table*}

\section{Experiments}
\label{sec:exp}

We evaluate our framework on two challenging simulation benchmarks—CALVIN~\citep{mees2022calvin} and MetaWorld~\citep{yu2019meta}, as well as across real-world environments involving diverse robotic manipulation tasks.
In this section, we first introduce our experimental setup, followed by evaluations on the three selected robotics environments, and finally ablation studies.

\subsection{Implementation Details}
Our \modelname comprises two components, \systwoname and \sysonename. 
We initialize our \systwoname from a pretrained latent diffusion model from ~\citep{rombach2022high, stable_diffusion_v1_5_huggingface} that has been trained on large-scale image-text datasets. 
The additional U-Net weights we use for our additional visual conditioning are zero-initialized to ensure that the pretrained network behavior is preserved at the beginning of training, and the model can gradually adapt to the additional input modality. We encode the language instruction and extract gripper view visual tokens using CLIP model. During the inference, we use 25 diffusion steps to generate the final pixel motion prediction.

Our \sysonename conditions on visual, textual, and robotic state modalities with different encoders for each input. The visual encoder is a pretrained ConvNeXt-S variant of DINOv3~\citep{simeoni2025dinov3}, and the text encoder is a T5-small pretrained model. 
% The state encoder are randomly initialized.

\subsection{CALVIN Experiments}
\label{subsec:calvin_exp}

\bhdr{Dataset:}
We first evaluate our \modelname on the CALVIN benchmark~\citep{mees2022calvin}. This simulated benchmark measures the long-horizon capability of robotic manipulation tasks. We focus on the most challenging ABC$\rightarrow$D task setting, where the model is trained on the A, B, and C environments and then evaluated in the unseen D environment. We follow standard evaluation protocol from ~\citep{Hu2024VideoPP, wu2023unleashing}. More details is provided in Supplementary Materials.

% Several prior works also report results using pretraining on external data, including \citep{Hu2024VideoPP,Ran25LTM, Zhang2025DreamVLAAV, Gu2023SeerLI}. We train our model under this setting as well, where we use the DROID dataset \citep{khazatsky2024droid} for our pretraining. 

% \bhdr{Evaluation:}
% We follow standard evaluation protocol from ~\citep{Hu2024VideoPP}, which evaluates a given policy on 1000 episodes each containing 5 continuous tasks (i.e. task $i$ starts from the end state of task $i-1$, which is often different to what is encountered in demonstrations within the training data). 
% For each task, at most 360 action steps are performed unless the task is successfully completed prior to that. The success rate for each consecutive task is averaged across the 1000 episodes and reported. Considering the 5 continuous tasks as a sequences, the average number of tasks completed by the policy (i.e. average length) is also reported.  

\bhdr{Results:} 
We report results under two training settings, first without using any external robotic demonstration data in \Cref{tbl:calvin_only} and second with external robotic demonstration data (DROID) in \Cref{tbl:calvin_plus}. 
Since we follow evaluation protocol identical to \citep{Hu2024VideoPP,Ran25LTM}, baseline results in our tables are directly borrowed from these prior works.    

In \Cref{tbl:calvin_only}, our \modelname achieves state-of-the-art results, highlighting the promise of pixel-motion based representations for complex robotic manipulation tasks. We also report results for our only the low-level \sysonename as Enhanced DP. These results highlight the clear impact of pixel motions in achieving the strong results of our overall \modelname framework. Example rollouts are presented in the appendix. % \Cref{fig:calvin_fs}.

In the case of using external robotic data, direct comparison to prior works (where different approaches use different pretraining data) is less straightforward. 
We report results for our \modelname that is trained jointly on the DROID dataset and CALVIN ABC$\rightarrow$D split.  
Our \modelname outperforms several recent works and performs competitively against SOTA methods VPP \citep{Hu2024VideoPP} and DreamVLA \citep{Zhang2025DreamVLAAV}. 
In \Cref{tbl:calvin_plus}, VPP benefits from significantly more videos (including 193k human manipulation trajectories, 179k robot manipulation trajectories, CALVIN,  MetaWorld, and additional real-world datasets) in its pretraining. Similarly, DreamVLA was first pretrained on a language-free split of the CALVIN and the full DROID dataset.

Overall, our \modelname achieves state-of-the-art performance on CALVIN benchmark, demonstrating the scalability as well as strong data efficiency of intermediate pixel-motion based VLA approaches.

\begin{table*}[t]

\centering
\small
\def\arraystretch{1.2}  %
\setlength\tabcolsep{1.2em}  %
\caption{\textbf{MetaWorld task success rate:}
Our \modelname achieves state-of-the-art performance on MetaWorld.} %

\scalebox{0.8}{%
\begin{tabular}{lcccccccccccc}
\toprule
 Method & \footnotesize \rotatebox{70}{door-open} & \footnotesize \rotatebox{70}{door-close} & \footnotesize \rotatebox{70}{basketball} & \footnotesize \rotatebox{70}{shelf-place} & \footnotesize \rotatebox{70}{btn-press} & \footnotesize \rotatebox{70}{btn-top} & \footnotesize \rotatebox{70}{faucet-close} & \footnotesize \rotatebox{70}{faucet-open} & \footnotesize \rotatebox{70}{handle-press} & \footnotesize \rotatebox{70}{hammer} & \footnotesize \rotatebox{70}{assembly} & \footnotesize \rotatebox{70}{Overall} \\
\midrule
BC-Scratch~\citep{nair2022r3m} & 21.3 & 36.0 & 0.0 & 0.0 & 34.7 & 12.0 & 18.7 & 17.3 & 37.3 & 0.0 & 1.3 & 16.2 \\
BC-R3M~\citep{nair2022r3m} & 1.3 & 58.7 & 0.0 & 0.0 & 36.0 & 4.0 & 18.7 & 22.7 & 28.0 & 0.0 & 0.0 & 15.4 \\
Diffusion Policy & 45.3 & 45.3 & 8.0 & 0.0 & 40.0 & 18.7 & 22.7 & 58.7 & 21.3 & 4.0 & 1.3 & 24.1 \\
UniPi~\citep{du2023unipi} (With Replan) & 0.0 & 36.0 & 0.0 & 0.0 & 6.7 & 0.0 & 4.0 & 9.3 & 13.3 & 4.0 & 0.0 & 6.1 \\
Im2Flow2Act~\citep{Xu2024FlowAT} & 0.0 & 0.0 & 0.0 & 4.0 & 6.3 & 0.0 & 7.3 & 4.7 & 0.0 & 0.0 & 0.0 & 2.0 \\
ATM~\citep{Wen2023AnypointTM} & 75.3 & 90.7 & 24.0 & 16.3 & 77.3 & 76.7 & 50.0 & 62.7 & 92.3 & 4.3 & 2.0 & 52.0 \\
AVDC~\citep{Ko2023LearningTA} (Flow) & 0.0 & 0.0 & 0.0 & 0.0 & 1.3 & 40.0 & 42.7 & 0.0 & 66.7 & 0.0 & 0.0 & 13.7 \\
AVDC~\citep{Ko2023LearningTA} (Default) & 72.0 & 89.3 & 37.3 & 18.7 & 60.0 & 24.0 & 53.3 & 24.0 & 81.3 & 8.0 & 6.7 & 43.1 \\
LTM~\citep{Ran25LTM} & 77.3 & 95.0 & 39.0 & 20.3 & 82.7 & 84.3 & 52.3 & 68.3 & 98.0 & 10.3 & 7.7 & 57.7 \\ \rowcolor{Gray}
DAWN (ours) & 94.7 & 97.3 & 42.0 & 24.7 & 92.0 & 91.7 & 76.3 & 79.0 & 98.0 & 12.7 & 10.7 & 65.4 \\ 
\bottomrule
\end{tabular}
}
\label{tbl:mw_eval}

\end{table*}

\subsection{Meta-World Experiments}
\label{subsec:mw_exp}

% We next evaluate \modelname on the MetaWorld~\citep{yu2019meta} simulated environment containing a Sawyer robot arm. 
% We focus on 11 challenging tasks constructed following \citep{Ko2023LearningTA,Ran25LTM} since the original benchmark is not language conditioned. 
% We select this environment-tasks setting to enable direct comparison to closely related prior works \citep{Ko2023LearningTA,Wen2023AnypointTM,Xu2024FlowAT,Ran25LTM} that also leverage pixel or point trajectories for robot manipulation tasks. 

\bhdr{Dataset:}
We next evaluate \modelname on the MetaWorld~\citep{yu2019meta} simulated environment containing a Sawyer robot arm. 
We focus on 11 challenging tasks constructed following \citep{Ko2023LearningTA,Ran25LTM} since the original benchmark is not language conditioned. 
We use the training split from \citep{Ko2023LearningTA,Ran25LTM} containing 165 actionless videos for \systwoname training and 220 task demonstrations across the 11 tasks for \sysonename training. 
All baseline experiments and results are identical to those reported in prior works AVDC \citep{Ko2023LearningTA} and LTM \citep{Ran25LTM}. 
% The baselines ``BC'' refer to behaviour cloning, with BC-Scratch containing a ResNet initialized from scratch and BC-R3M using a ResNet initialized from R3M \citep{nair2022r3m}. All baselines are trained as described in \citep{Ko2023LearningTA,Ran25LTM}.

\bhdr{Results:}
We report these results in \Cref{tbl:mw_eval}. 
Our approach achieves clear performance improvements compared to prior works on this challenging benchmark. We particularly emphasize the improved \textit{semantic understanding} of our \modelname framework: notice how \modelname achieves significantly better performance on visually similar but semantically dissimilar task pairs such as \texttt{open-door} vs \texttt{close-door}. We attribute this to our efficient latent diffusion formulation that enables scalable language-video pretraining, which in turn endows our model with stronger language understanding.  
We also highlight the improved performance in tasks such as \texttt{basketball} and \texttt{assembly}, which we attribute to our action-expert design choices that enable better robot state awareness. 

% We take these results as clear indication to how our design choices elevate the capabilities of intermediate pixel motion based VLA approaches, establishing the promise of this direction. 

\begin{table*}[t]
\centering
\small
\scriptsize
\caption{\textbf{Real-world single lift-and-place evaluation}. For each task, we evaluate 20 episodes with random initialization position. }

\resizebox{\textwidth}{!}{%
\begin{tabular}{l|cc|cc|cc|cc|cc|cc|c}
\toprule
& \multicolumn{2}{c|}{\textbf{Apple}} 
& \multicolumn{2}{c|}{\textbf{Avocado}} 
& \multicolumn{2}{c|}{\textbf{Banana}} 
& \multicolumn{2}{c|}{\textbf{Grape}} 
& \multicolumn{2}{c|}{\textbf{Kiwi}} 
& \multicolumn{2}{c|}{\textbf{Orange}} 
& \multicolumn{1}{c}{\textbf{Inference}} \\
& Success $\uparrow$ & Wrong  Obj. $\downarrow$ & Success $\uparrow$ & Wrong  Obj. $\downarrow$ & Success $\uparrow$ & Wrong  Obj. $\downarrow$ & Success $\uparrow$ & Wrong  Obj. $\downarrow$ & Success $\uparrow$ & Wrong  Obj. $\downarrow$ & Success $\uparrow$ & Wrong  Obj. $\downarrow$ & \multicolumn{1}{c}{\textbf{Efficiency (ms)} $\downarrow$} \\
& (i)$\rightarrow$(ii) & (iii)$\rightarrow$(iv) & (i)$\rightarrow$(ii) & (iii)$\rightarrow$(iv) &
(i)$\rightarrow$(ii) & (iii)$\rightarrow$(iv) &
(i)$\rightarrow$(ii) & (iii)$\rightarrow$(iv) &
(i)$\rightarrow$(ii) & (iii)$\rightarrow$(iv) &
(i)$\rightarrow$(ii) & (iii)$\rightarrow$(iv) & \\
\midrule
Enhanced DP
& 5$\rightarrow$4 & 9$\rightarrow$8 
& 6$\rightarrow$6 & 6$\rightarrow$4 
& 5$\rightarrow$4 & 6$\rightarrow$4 
& 4$\rightarrow$3 & 8$\rightarrow$6 
& 5$\rightarrow$5 & 6$\rightarrow$5 
& 4$\rightarrow$4 & 8$\rightarrow$7 
& 112.77 \\

$\pi_0$~\citep{black2024pi_0} 
& 10$\rightarrow$9 & 9$\rightarrow$9 
& 6$\rightarrow$6 & 12$\rightarrow$10 
& 5$\rightarrow$3 & 11$\rightarrow$6 
& 8$\rightarrow$5 & 10$\rightarrow$8 
& 5$\rightarrow$3 & 12$\rightarrow$12 
& 8$\rightarrow$7 & 11$\rightarrow$11 
& 571.89 \\

% \modelname
% & \textbf{12$\rightarrow$10} & \textbf{2$\rightarrow$2} 
% & \textbf{10$\rightarrow$10} & \textbf{1$\rightarrow$1} 
% & \textbf{9$\rightarrow$7} & \textbf{0$\rightarrow$0} 
% & \textbf{15$\rightarrow$10} & \textbf{1$\rightarrow$0} 
% & \textbf{13$\rightarrow$11} & \textbf{0$\rightarrow$0} 
% & \textbf{11$\rightarrow$11} & \textbf{2$\rightarrow$2} 
% & 988.44 \\
% \midrule 

VPP~\citep{Hu2024VideoPP}
& {16$\rightarrow$14} & {2$\rightarrow$2} 
& {15$\rightarrow$15} & {2$\rightarrow$0} 
& {15$\rightarrow$14} & {0$\rightarrow$0} 
& {17$\rightarrow$17} & {1$\rightarrow$0} 
& {15$\rightarrow$15} & {2$\rightarrow$0} 
& {16$\rightarrow$14} & {0$\rightarrow$0} 
& 190.55 \\

\modelname
& \textbf{19$\rightarrow$19} & 0$\rightarrow$0 
& \textbf{20$\rightarrow$19} & 0$\rightarrow$0
& \textbf{17$\rightarrow$16} & 0$\rightarrow$0
& \textbf{19$\rightarrow$19} & 0$\rightarrow$0
& \textbf{17$\rightarrow$16} & 2$\rightarrow$2
& \textbf{18$\rightarrow$16} & 0$\rightarrow$0 
& 319.82 \\
\bottomrule
\end{tabular}
}
\label{tbl:rw_sr}
\end{table*}

\begin{figure}
    \centering
    \scalebox{0.9}{
    \includegraphics[width=\columnwidth]{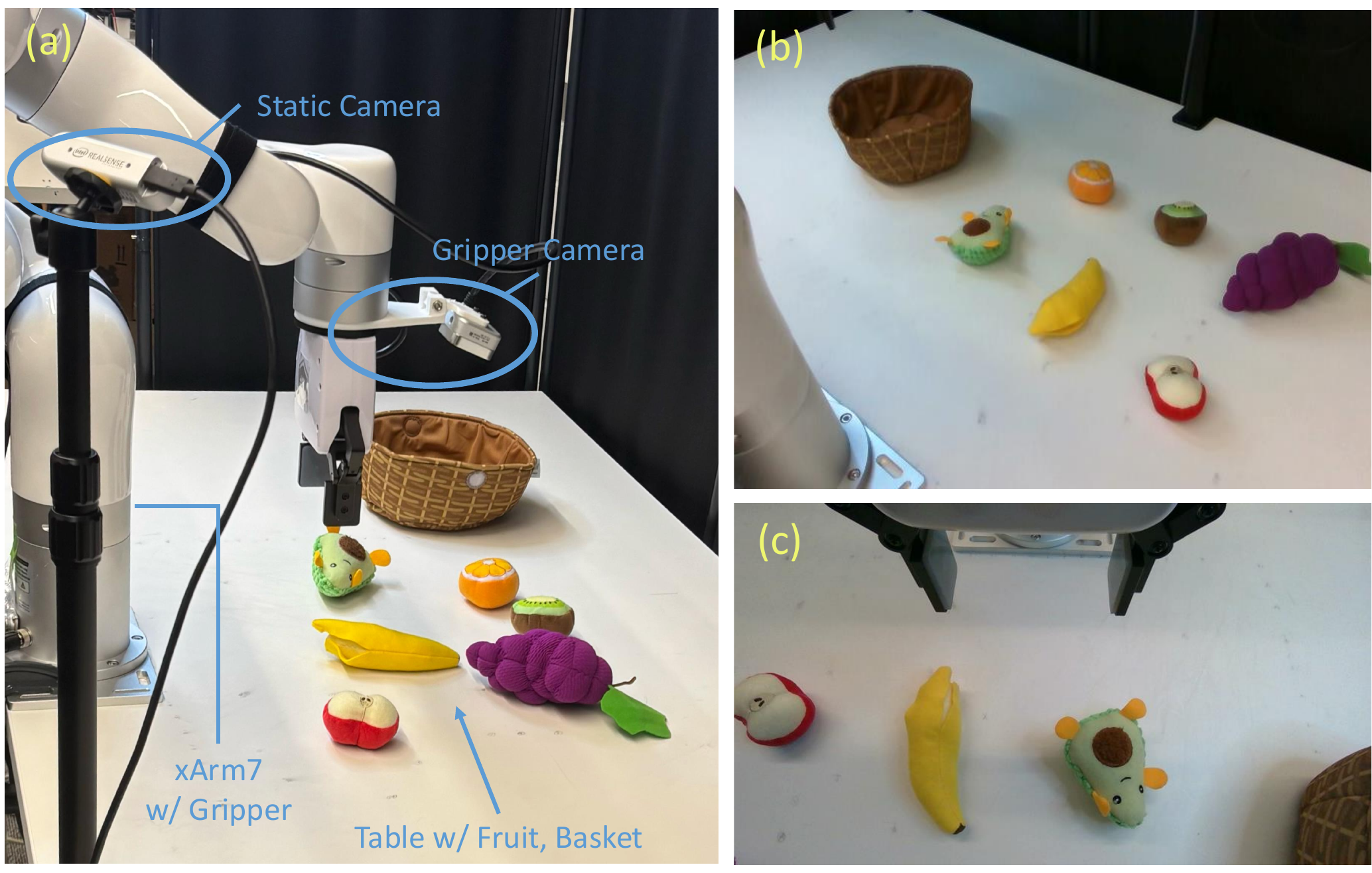}
    }
    \captionof{figure}{\textbf{Real-world environment examples.} a) Our single-arm environment includes a robot arm and two cameras. They are stereo RGB cameras, but we only use one RGB view from each camera. b) The RGB image from the static camera. c) The RGB image from the gripper camera.}
    \label{fig:rw_set}
    \vspace{-5mm}
\end{figure}

\subsection{Real-world Single-arm Manipulation}
\label{subsec:rw_exp}
We set up our single-arm environment with a 7-DoF xArm7 robot arm and two RGB cameras: one providing a fixed third-person view from the right side of the arm, and the other mounted above the gripper (see \Cref{fig:rw_set}). A dataset of 1000 episodes is then collected, comprising lift-and-place manipulations involving six types of toys and a container.

\bhdr{Implementation:} We compare our approach against three strong baselines. The first is our modification of Diffusion Policy~\citep{Chi2023DiffusionPV}, Enhanced DP, which is identical to our \sysonename but without pixel motion from a \systwoname. This model is pretrained on CALVIN ABC dataset. The second baseline is $\pi_0$~\citep{black2024pi_0}, where we initialize from the $\pi_0$ base model and apply Low-Rank Adaptation (LoRA)~\citep{hu2022lora}. 
The third is VPP \cite{Hu2024VideoPP}, initialized from their official pretrained checkpoint. 
All methods are fine-tuned on our collected real-world dataset for 100k steps. The task is highly challenging for the policy to learn, as a total of only 1k episodes across 12 tasks provides very limited training data.

\begin{figure*}[t]
    \centering
    \vspace{-3mm}
    \includegraphics[width=0.95\textwidth]{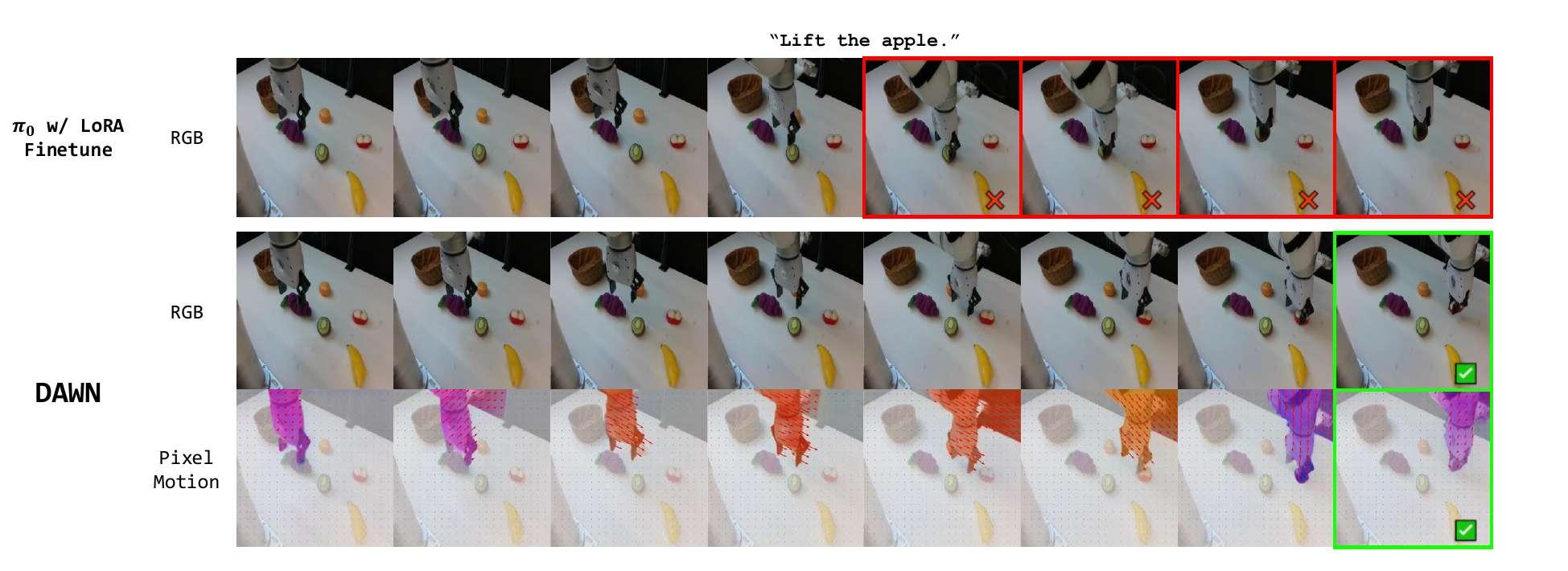}
    
    \caption{\textbf{Real world rollout examples}. For “lift the apple,” $\pi_0$ lifts the wrong object (top), while DAWN succeeds in the same setting (middle). Pixel motion predicted by \systwoname is shown in the last row. }

    \label{fig:rw_fs}
\end{figure*}

\begin{figure*}[t]
    \centering
    \includegraphics[width=0.89\textwidth]{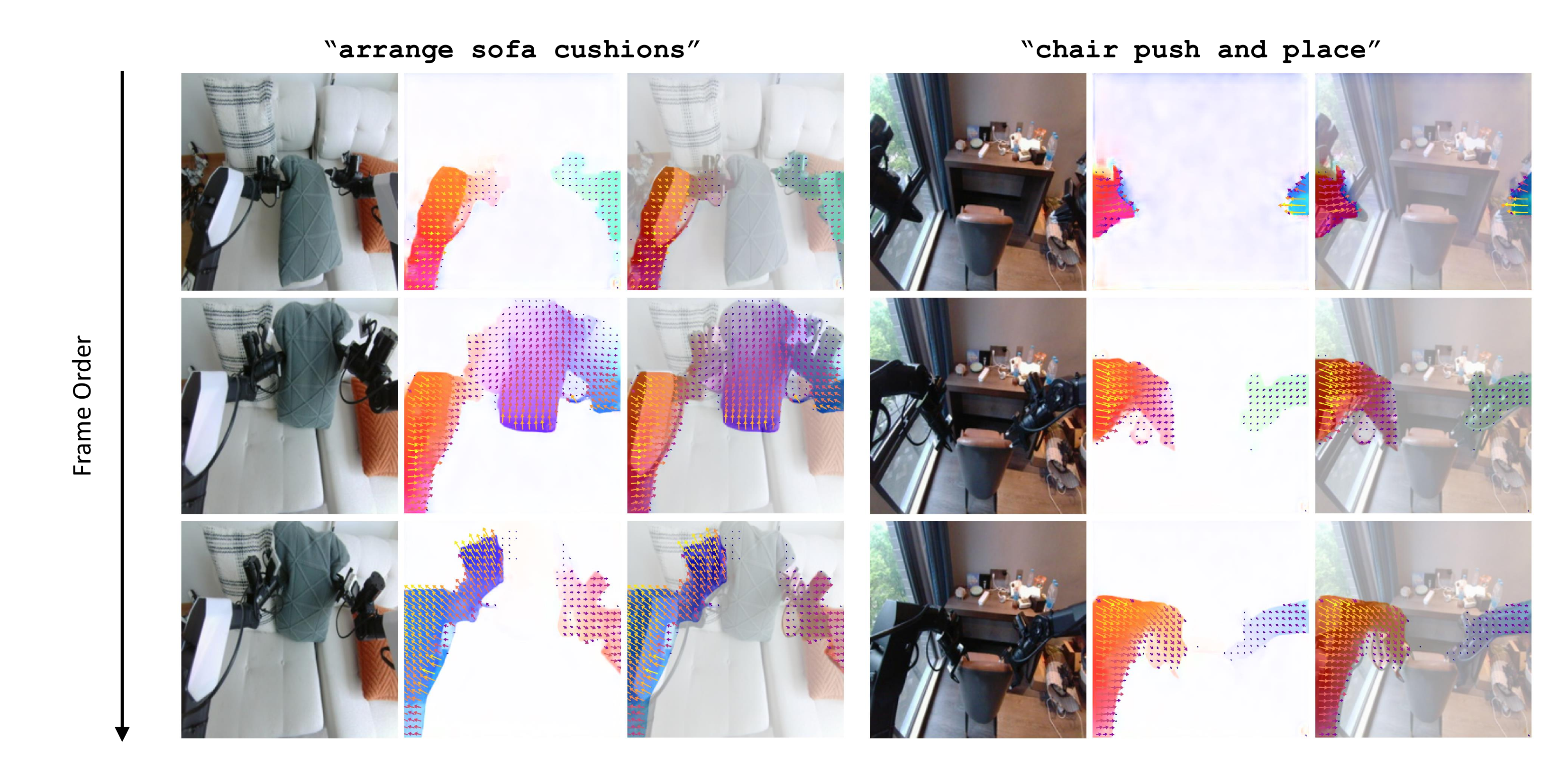}
    \vspace{-3mm}
    \caption{\textbf{Galaxea pixel motion prediction examples}. The first column shows the one test image sequence given the task of ``arrange sofa cushions''. The second column shows the test image sequence given the task of ``chair push and place''. Each group shows the original head-camera observation and the visualizations of corresponding pixel motions predicted by \systwoname.}
    \label{fig:rw_galaxea}
\end{figure*}

\bhdr{Evaluation:} We evaluate all methods using the lift-and-place task pair with different objects, where the robot is instructed to lift a specified object and place it into a container. We record the number of episodes in which the robot: (i) successfully lifts the correct object, (ii) successfully places the correct object, (iii) lifts an incorrect object at the end with 500 max steps, and (iv) places the incorrect object from the previous lifting. %
Note that (iii) and (iv) are still classified as failures, though they differ from complete failures to grasp or place the object. We include these cases to provide a clearer understanding of the failure patterns.
Each episode begins from a randomly initialized environment, and we run 20 episodes per task in total.

\bhdr{Results:} Table~\ref{tbl:rw_sr} shows that \modelname achieves the highest success rates across nearly all object categories, outperforming $\pi_0$, VPP, and Enhanced DP.  
Compared to $\pi_0$ and VPP, \modelname exhibits better semantic grounding and more consistent lifting of the correct object. \Cref{fig:rw_fs} illustrates a representative failure case for $\pi_0$, contrasted with a successful trajectory from \modelname. The Enhanced DP baseline (without pixel motion) frequently grasps the wrong object, demonstrating that structured pixel motion from \systwoname provides complementary benefits and can further strengthen even strong two-stage diffusion based methods such as VPP.

% Results in \Cref{tbl:rw_sr} demonstrate that our method achieves higher success in lifting and placing the correct object compared to the baselines, despite using far fewer parameters than $\pi_0$. In contrast, without the pixel motion provided by \systwoname, our Enhanced Diffusion Policy baseline frequently fails by lifting the wrong object or completely failing the task.
% \modelname is both more accurate and more parameter-efficient than the baselines in this set of challenging tasks. 

% Compared to  $\pi_0$ and VPP, \modelname consistently outperforms and exhibits better semantic awareness, lifting the correct object more often. This demonstrates that structured pixel motion provides complementary benefits and can further strengthen even strong two-stage diffusion based methods such as VPP.
% \Cref{fig:rw_fs} shows an example episode where $\pi_0$ fails to follow the instruction and lifts the wrong object, and \modelname can lift the correct object.

\bhdr{Efficiency:} 
We report the average inference runtime for each chunk of action steps in the last column of \Cref{tbl:rw_sr}. 
Enhanced DP is the fastest baseline since it uses a single diffusion model without any additional high-level module. 
VPP is slightly slower but remains efficient because its video diffusion model operates in a single forward pass to extract predictive features, without explicitly generating video frames or pixel-level outputs. 
\modelname has higher computational time due to its two-stage design: the Motion Director first predicts explicit pixel motion and the Action Expert then produces actions conditioned on this motion. 
Despite the  overhead, \modelname still runs at practical closed-loop frequencies and achieves the highest overall success rates, indicating that the benefits of explicit motion reasoning outweigh the added inference time.
% Even so, \modelname is still faster and substantially more accurate than $\pi_{0}$. 

% We measure the average runtime of each end-to-end step to evaluate computational efficiency, as shown in the last column of \Cref{tbl:rw_sr}. Enhanced DP achieves the highest efficiency by employing only single diffusion policy model. \modelname exhibits slower inference due to its two-stage design that explicitly propagates pixel motion between \sysonename and \systwoname. However, the enhanced variant \modelB adopts an implicit motion representation, significantly reducing computational overhead. This design improves the inference efficiency to - faster than $\pi_0$, comparable to VPP, while maintaining the best overall task performance. 

% \begin{wraptable}[9]{r}{0.45\textwidth}
\begin{table}
\centering
\vspace{-1em}
\begin{tabular}{l|c}
\toprule
Method                    & MSE$\downarrow$   \\ \midrule
Enhanced Diffusion Policy & 0.128 \\
Ours                      & 0.117 \\ \bottomrule
\end{tabular}
\caption{\textbf{Bimanual manipulation results.} Mean squared error (MSE) of action prediction on the test set. Our approach achieves lower error compared to the baseline.}
\label{tab:bimanual}
\end{table}

\subsection{Bimanual Manipulation}
\label{subsec:bimanual}
We further extend our framework in a more challenging bimanual setting using the Galaxea R1-Lite. The setup includes two 7-DoF arms with a gripper, a head static camera, and two wrist-mounted cameras. Similarly, though the cameras are stereo cameras, we only use a single RGB frame from each camera for our experiments.

\bhdr{Implementation and Evaluation:} We train our model on the subset of Galaxea Open-World Dataset~\cite{jiang2025galaxea}. The subset contains more than 100 hours of real-world demos, covering nine different types of challenging bimanual tasks in seven real-world environments.

We report the mean squared error (MSE) of predicted joint actions on a test set of 300 episodes. We compare our proposed model against the baseline Enhanced Diffusion Policy (without pixel motion from \systwoname), as shown in ~\Cref{tab:bimanual}. Results show that our approach achieves consistently lower MSE, indicating that structured pixel motion improves the accuracy of action prediction even in the more complex bimanual setting. \Cref{fig:rw_galaxea} shows two example samples from the test set where \modelname can predict the correct pixel motion corresponding to the given observations. These findings suggest that the benefits of our two-stage framework generalize beyond single-arm tasks and extend to multi-arm coordination scenarios.

\begin{table}[t]
\centering

    \vspace{-5mm}
    \caption{
     Ablation study on CALVIN dataset.
     }
     
    \scalebox{0.7}{
        \begin{tabular}{@{}lc@{}}
        \toprule
        \multicolumn{1}{l|}{Setting}                             & Avg. Length \\ \midrule
        \multicolumn{2}{l}{\textit{(a) Pixel Motion vs RGB Goal}}                          \\ \midrule
        \multicolumn{1}{l|}{None}                                & 2.78     \\
        \multicolumn{1}{l|}{RGB Goal}                             & 3.21     \\
        \multicolumn{1}{l|}{Pixel Motion w/o pretrained}                             & 3.42     \\
        \rowcolor{Gray}
        \multicolumn{1}{l|}{Pixel Motion}                         & \textbf{4.00}    \\ \midrule
        \multicolumn{2}{l}{\textit{(b) Gripper View}}                          \\ \midrule
        \multicolumn{1}{l|}{VPP w/o gripper view}                                & 3.58     \\
        \multicolumn{1}{l|}{\modelname w/o Gripper view}                                & 3.74     \\
        
        \rowcolor{Gray}
        \multicolumn{1}{l|}{\modelname w/ Gripper View}                             & \textbf{4.00}     \\ \midrule
        
        \multicolumn{2}{l}{\textit{(c) \# of Diffusion steps of \systwoname}}                          \\ 
        \midrule
        \multicolumn{1}{l|}{2}                                & 3.88     \\
        \multicolumn{1}{l|}{10}                             & 3.96     \\
        \rowcolor{Gray}
        \multicolumn{1}{l|}{25}                         & \textbf{4.00}    \\ 
        \multicolumn{1}{l|}{40}                             & 3.95     \\

        \bottomrule
        
        \end{tabular}
    }  
    \vspace{-5mm}
    \label{tab:ablation_full}
\end{table}

\subsection{Ablation Study}
\label{subsec:ablate}

We conduct ablation experiments on the CALVIN ABC$\rightarrow$D benchmark to assess the impact of (a) structured pixel motion representation, (b) gripper view conditioning, and (c) the number of diffusion steps (See \Cref{tab:ablation_full}).

\bhdr{(a) Pixel Motion, RGB Goal, and pretraining.}
We compare two variants against our default setting: (i) RGB goal image conditioning instead of pixel motion; (ii) Only \sysonename w/o pixel motion, and (iii) generating pixel motion with a denoising U-Net trained from scratch. 
As shown in \Cref{tab:ablation_full} (a), pixel motion yields the best performance, highlighting its utility as a structured and interpretable intermediate, and our method benefits a lot from the pretrained image generation model, even though the model was not trained for pixel motion generation before.

\bhdr{(b) Gripper View Conditioning.}
We further ablate the effect of adding egocentric gripper-mounted observation to \systwoname. Removing the gripper view leads to performance degradation (3.74 vs. 4.00), while prior methods such as VPP degrade further (3.58). These results confirm that the additional viewpoint facilitates reasoning about occlusions and fine-grained hand-object interactions.

\bhdr{(c) Diffusion Steps of \systwoname.}
\systwoname module can capture meaningful motion information even at 2 diffusion steps (3.88). Increasing the number of steps steadily improves performance, peaking at 25 (4.00) and can't gain more beyond that (e.g., 40 steps with 3.95).

\section{Conclusion}
\label{sec:conc}

In this work, we present a two-stage diffusion-based visuomotor framework for robot manipulation, termed \modelname, which achieves state-of-the-art performance on CALVIN, MetaWorld, and real-world benchmarks.  Instead of using manifest visual information in RGB space, we use explicit dense pixel motion representations as a structured interface between a latent diffusion \systwoname and a diffusion-based \sysonename. This design bridges hierarchical motion decomposition and end-to-end agents while preserving interpretability and modularity. By instantiating both stages with modern diffusion models and leveraging strong pre-trained vision language backbones, \modelname delivers high data efficiency and robust transfer, indicating that much of the gap between multi-stage tracking pipelines and VLA/latent-feature hierarchies stems not from the framework itself but from underpowered high- and low-level components. We hope \modelname encourages re-examining structured intermediate representations as a practical path to interpretable, data-efficient robot control.

\section*{Acknoledgements}
This research was financially supported by the Ministry of Trade, Industry, and Energy (MOTIE), Korea, under the ``Global Industrial Technology Cooperation Center(GITCC) program" supervised by the Korea Institute for Advancement of Technology (KIAT).(Task No. P0028420). This research was supported by the National Research Council of Science \& Technology(NST) grant by the Korea government(MSIT) (No. GTL25041-000).

{
    \small
    \bibliographystyle{ieeenat_fullname}
    \bibliography{main}
}

\clearpage
\setcounter{section}{0}
\maketitlesupplementary

\section{Training Details} 
We train all models on a single node with 4 NVIDIA A6000 GPUs. 
For \systwoname, we train for 100k iterations with a per-GPU batch size of 16.
For \sysonename, we train for 10k iterations with a per-GPU batch size of 64.
We use the AdamW optimizer with a learning rate of $1 \times 10^{-4}$.
Mixed precision training is used to reduce memory usage and improve throughput. 
All training is implemented in PyTorch with the HuggingFace Diffusers and Transformers libraries. 

% In our approach, we adopt a hierarchical inference strategy where \systwoname predicts a pixel motion plan, and \sysonename executes this plan by directly applying 10 consecutive low-level action steps before requesting a new pixel motion. This design reduces the computational overhead of repeatedly invoking the diffusion-based planner, while ensuring that each high-level motion is translated into a temporally coherent sequence of actions.

\section{Dataset details}
\subsection{CALVIN}
\bhdr{Dataset:} CALVIN is an open source simulated benchmark to learn long-horizon language-conditioned tasks, which contains 4 different simulation environments-A, B, C, D. While each split (A–D) shares the same robotic setup, variations in object placement, textures, lighting, and distractors ensure that models cannot rely on memorization but must instead demonstrate robust visuomotor understanding. The 34 manipulation tasks span a wide range of skills such as pushing, placing, rotating, toggling switches, and opening drawers, all expressed through natural language instructions.

\bhdr{Evaluation:}
We follow standard evaluation protocol from ~\citep{Hu2024VideoPP}, which evaluates a given policy on 1000 episodes each containing 5 continuous tasks (i.e. task $i$ starts from the end state of task $i-1$, which is often different to what is encountered in demonstrations within the training data). For each task, at most 360 action steps are performed unless the task is successfully completed prior to that. The success rate for each consecutive task is averaged across the 1000 episodes and reported. Considering the 5 continuous tasks as a sequences, the average number of tasks completed by the policy (i.e. average length) is also reported.  

\subsection{DROID}
DROID is a large-scale ``in-the-wild'' robot manipulation dataset featuring 76k real demonstration trajectories across 564 varied scenes and 86 tasks. It provides over 350 hours of interaction data, with diverse viewpoints, object types, and natural instruction annotations.

\subsection{Real World}
\subsubsection{xArm Environment}
We constructed a dataset specifically for fine-tuning and real-world evaluation. The experimental platform consists of a 7-DoF xArm7 manipulator and two RGB cameras. An Intel RealSense D435 was positioned laterally to provide a third-person view of the workspace, while an Intel RealSense D405 was mounted above the gripper to capture a close-up view of the end-effector and its interactions with objects. Though both cameras are stereo cameras, we only use a single RGB view from each camera in all the experiments. This dual-camera setup enables complementary perspectives, facilitating both scene-level and fine-grained observations.

Data was collected through a leader–follower teleoperation scheme, where a human operator controlled a leader device to guide the motions of the xArm7 (follower). Each demonstration episode was restricted to a single atomic task, such as lifting a fruit, transporting it, or placing it into a basket. Episodes were initialized either from randomized joint configurations or from the terminal state of the preceding task, ensuring diversity in initial conditions. To further increase variability and promote generalization, we occasionally re-dropped and re-grasped objects within the same episode.

The resulting dataset comprises 1,000 episodes, with a minimum of 100 demonstrations allocated to each distinct task. This distribution ensures both task balance and sufficient coverage for downstream fine-tuning. Overall, the dataset provides a structured yet diverse collection of manipulation trajectories suitable for evaluating task-specific policies under realistic conditions.

\subsubsection{Galaxea R1 Lite Environment}
In order to evaluate on more complicated environment, we also constructed a dataset for fine-tuning and real-world evaluation. This environment contains a bi-manual Galaxea R1 Lite platform and three RGB cameras. A RGB camera was positioned medially to provide the head view of the workspace, while the other two Intel Realsense D405 was mounted above two grippers in each side to capture the close-up view of the end-effector and their interactions with objects. Though both of the gripper-view cameras are stereo cameras, we only use a single RGB view from each camera in all the experiments.

We set up a table top manipulation environment, which has two plush toys: a lion and an elephant, also a toy pot. In terms of the tasks, we are manipulating plush toys and the lid cover of the toy pot (one hand manipulates a plush toy, and the other hand helps open the lid cover of the toy pot). See \Cref{fig:rw_gr_fs} for the example and \Cref{tab:task_definitions} for the task description. 

\begin{table*}[t]
\centering
\begin{tabular}{ll}
\hline
\textbf{Task} & \textbf{Language} \\
\hline
left\_place\_lion &
left arm lift and place lion into pot, right arm helps open and close lid. \\

left\_place\_elephant &
left arm lift and place elephant into pot, right arm helps open and close lid. \\

left\_pick\_lion &
left arm pick lion from pot, right arm helps open and close lid. \\

left\_pick\_elephant &
left arm pick elephant from pot, right arm helps open and close lid. \\

right\_place\_lion &
right arm lift and place lion into pot, left arm helps open and close lid. \\

right\_place\_elephant &
right arm lift and place elephant into pot, left arm helps open and close lid. \\

right\_pick\_lion &
right arm pick lion from pot, left arm helps open and close lid. \\

right\_pick\_elephant &
right arm pick elephant from pot, left arm helps open and close lid. \\
\hline
\end{tabular}
\caption{Task definitions used in our Galaxea Real World experiments.}
\label{tab:task_definitions}
\end{table*}

We collected our data through a leader-follower teleoperation scheme, which human operates the Galaxea R1 Lite Teleop to guide the motions of the Galaxea R1 Lite (follower). Each demonstration episode was a single bi-manual pick-place task, such as picking the lion into the pot and helping to open the lid. The collected dataset comprises 150 episodes, which provides a structured collection of learnable bi-manual manipulation trajectories.

\section{More analysis}
\subsection{Computational trade-off}
As shown in Table~\ref{tab:latency}, reducing the number of reverse diffusion steps in the Motion Director during inference substantially lowers the latency, with performance degrading slightly. This demonstrates a practical and potential mechanism to support higher-frequency control by trading off motion fidelity for speed.
\begin{table}[h]
    \centering
    \small
    \def\arraystretch{1.0}  % height
    \setlength\tabcolsep{0.4em}  % width
    \scalebox{0.99}{
    \begin{tabular}{c|c|c}
    \hline
    \textbf{Diffusion Steps} & \textbf{Latency(ms)} & \textbf{Avg. Len} \\
    \hline
    % Enhanced DP & 27 & 2.78 \\ 
    2  & 32   & 3.88 \\
    10 & 60 & 3.96 \\ \rowcolor{Gray}
    25 & 99   & \textbf{4.00} \\
    40 & 149 & 3.95 \\
    \hline
    \end{tabular}
    }
    \vspace{-2mm}
    \caption{Trade-off between reverse diffusion steps on Motion Director and performance on CALVIN validation dataset.} % Fewer steps offer lower latency with modest performance degradation. We measure the average latency(ms) for each action step during inference under CALVIN environment.}
    \label{tab:latency}
\end{table}

\subsection{Motion Director metrics}
We include a quantitative comparison of pixel motion prediction error (MSE) against ground-truth optical flow between DAWN and LTM (\Cref{fig:of_metrics}). DAWN achieves lower prediction error, indicating much more accurate pixel motion quality. In addition, Table~6 (a) in the paper shows that using pixel motion consistently outperforms RGB or no-motion variants. 
\begin{table}[h]
    \centering
    \small
    \def\arraystretch{1.0}  % height
    \setlength\tabcolsep{0.4em}  % width
       \scalebox{0.99}{
    \begin{tabular}{c|c}
    \hline
    \textbf{Model} & \textbf{Error} \\
    \hline
    LTM  & $5.19 \times 10^{-4}$ \\ 
    DAWN & $1.23 \times 10^{-4}$\\
    \hline
    \end{tabular}
    }
    \vspace{-2mm}
    \caption{Motion Director pixel motion prediction error (MSE) on CALVIN validation dataset.}
    % \label{tab:hist}
    \label{fig:of_metrics}
\end{table}

\subsection{Ablation study on temporal offset $k$}
Table~\ref{tab:k} shows that performance is stable across different $k$, with $k{=}20$ yielding the best results and we use it throughout all our experiments.

\begin{table}[h]
       \centering
       \vspace{-2mm}
        \small
           \scalebox{0.99}{
           \begin{tabular}{c|cccc}
    \hline
    \textbf{$k$}        & 5    & 10   & 20 & 30   \\
    \hline
    \textbf{Avg. Len}   & 3.72 & 3.93 & \textbf{4.00}        & 3.67 \\
    \hline
    \end{tabular}
        }
        \vspace{-1em}
        \caption{Ablation study of different temporal offset $k$ in CALVIN environment.
        % Sensitivity to $K$ with fixed $K$ during training. Performance is stable around $K=20$, which we use in all experiments.
        }
        \label{tab:k}
\end{table}

\section{Qualitative Results}
This section includes a series of visualizations demonstrating how \modelname generates and executes pixel-motion plans across diverse environments. All videos and overlays are packaged locally and can be viewed through the provided index.html together with this supplementary.

\subsection{Bi-manual Pixel Motion Predictions}
\begin{figure*}[h]
    \centering
    \includegraphics[width=0.9\textwidth]{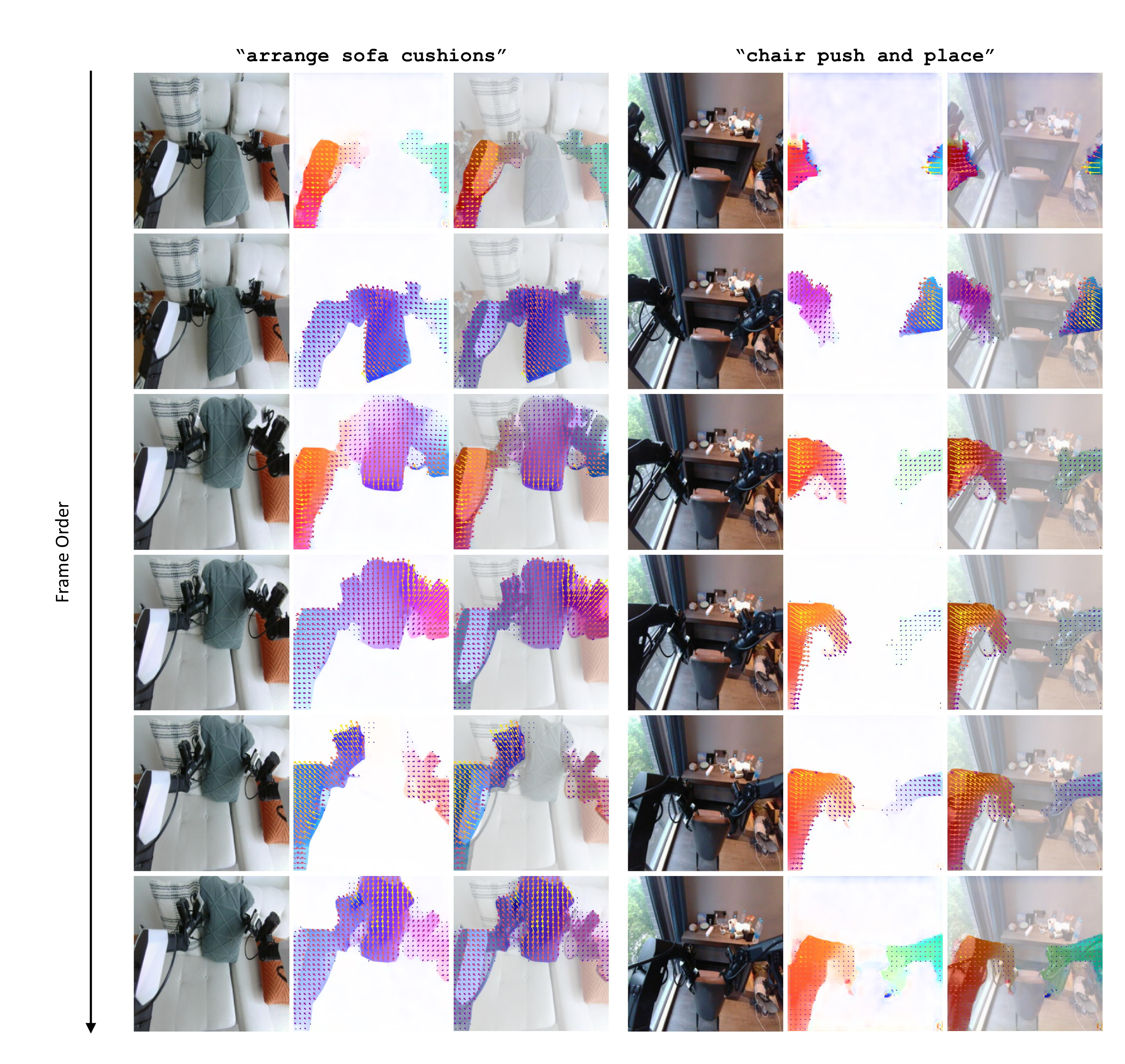}
    \vspace{-1.5em}
    \caption{\textbf{Galaxea-Open-World-Dataset pixel motion prediction examples}. The first column shows the one test image sequence given the task of ``arrange sofa cushions''. The second column shows the test image sequence given the task of ``chair push and place''. Each group shows the original head-camera observation and the visualizations of corresponding pixel motions predicted by \systwoname.} 
    \label{fig:gr_fs}
\end{figure*}

\begin{figure*}[h]
    \centering
    \includegraphics[width=0.9\textwidth]{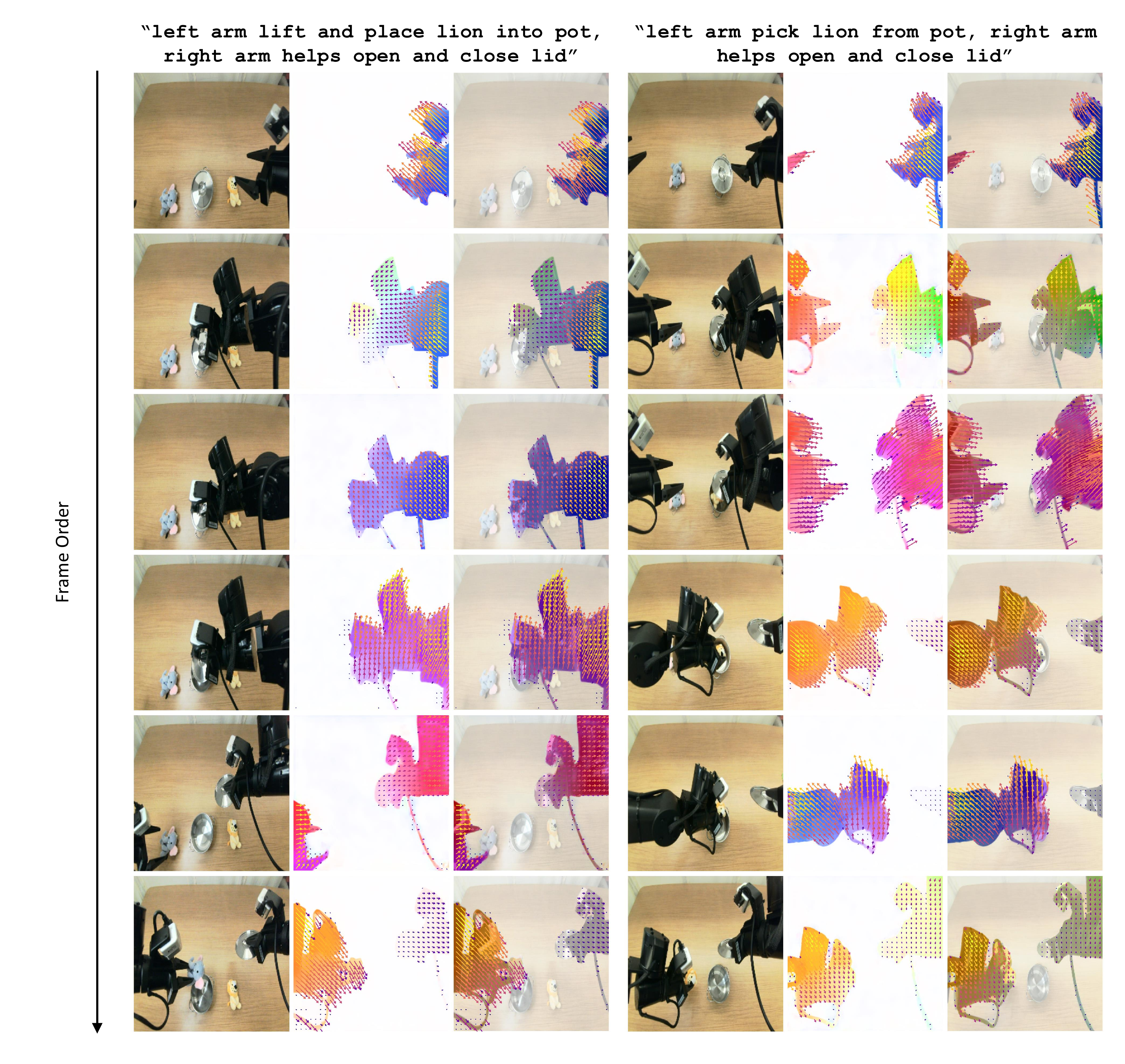}
    \vspace{-1.5em}
    \caption{\textbf{Real-World Galaxea pixel motion prediction examples}. The first column shows the one test image sequence given the task of ``left arm lift and place lion into pot, right arm helps open and close lid''. The second column shows the test image sequence given the task of ``left arm pick lion from pot, right arm helps open and close lid''. Each group shows the original head-camera observation and the visualizations of corresponding pixel motions predicted by \systwoname.} 
    \label{fig:rw_gr_fs}
\end{figure*}

We first showcase \systwoname's pixel-motion predictions on bimanual manipulation sequences. These include both our own recorded bimanual setup and Galaxea-Open-World-Dataset videos. In each case, we overlay the predicted pixel motion on each frame to reveal how the \systwoname captures coordinated left–right arm movements, object-relative displacements, and long-range motion cues. These examples highlight that the motion plans remain consistent even in visually complex or asymmetric dual-arm settings. Our two examples of Galaxea-Open-World-Dataset pixel motion prediction are presented in \Cref{fig:gr_fs}. Another two examples of our real world environment pixel motion prediction are presented in \Cref{fig:rw_gr_fs}.
\subsection{Real-World xArm Manipulation}
\begin{figure*}[h]
    \centering
    \includegraphics[width=0.7\textwidth]{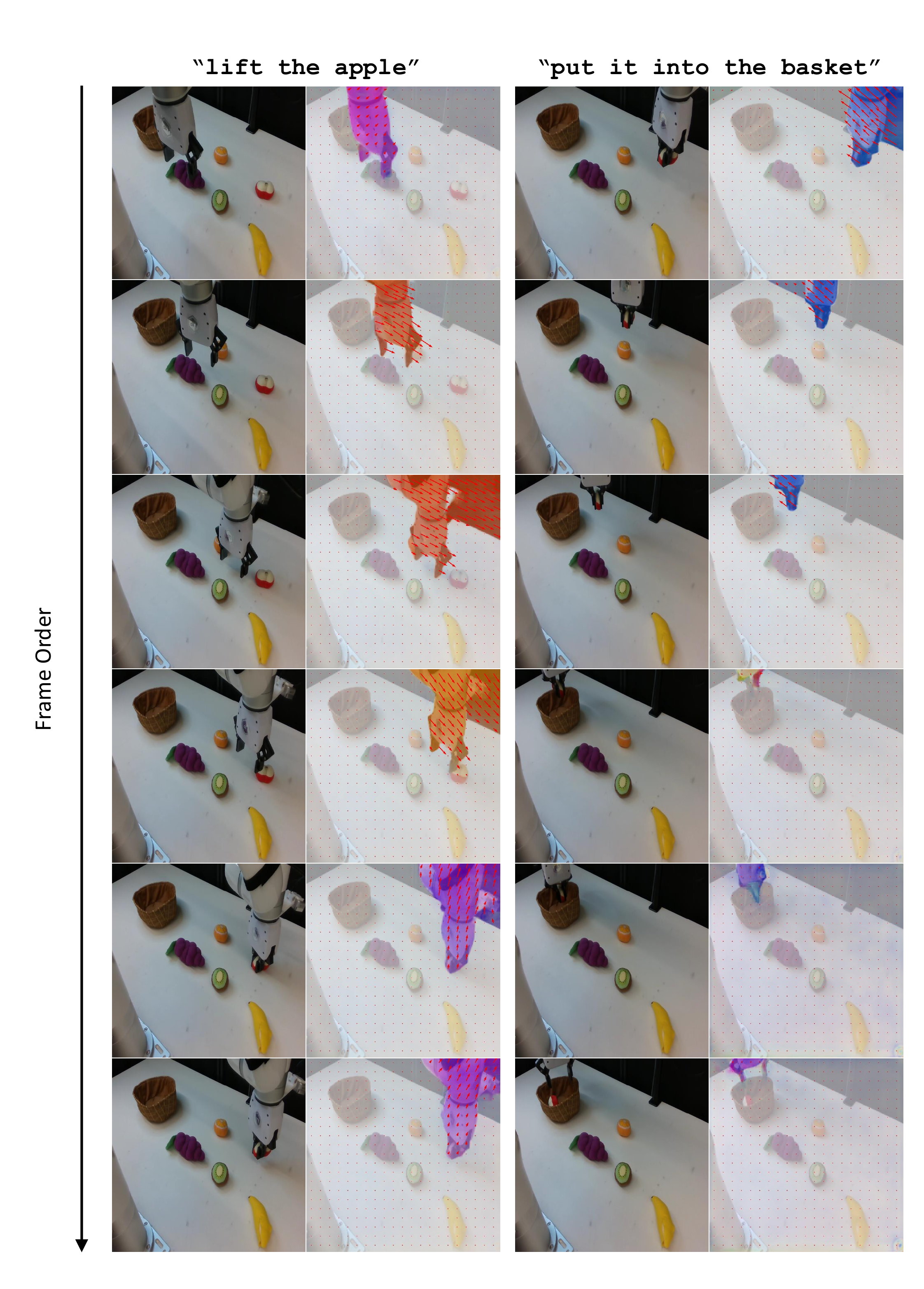}
    \vspace{-1.5em}
    \caption{\textbf{xArm rollout examples}. The first column shows the observation sequence given the task of ``lift the apple''. The second column shows the observation sequence given the task of ``put it into the basket''. Each group shows the original static-camera observation and the visualizations of corresponding pixel motions predicted by \systwoname.} 
    \label{fig:xarm_fs}
\end{figure*}

Next, we provide full rollout videos from our real-world xArm7 platform (see \Cref{fig:xarm_fs}). For every rollout, we include third-person view and frame-by-frame pixel-motion overlays. These visualizations show that the robot’s actual behavior reliably follows the predicted motion. This makes the high-level plan interpretable, which is one of the key advantages of using structured pixel-motion as the intermediate representation.

\begin{figure*}[h]
    \centering
    \includegraphics[width=1.05\textwidth]{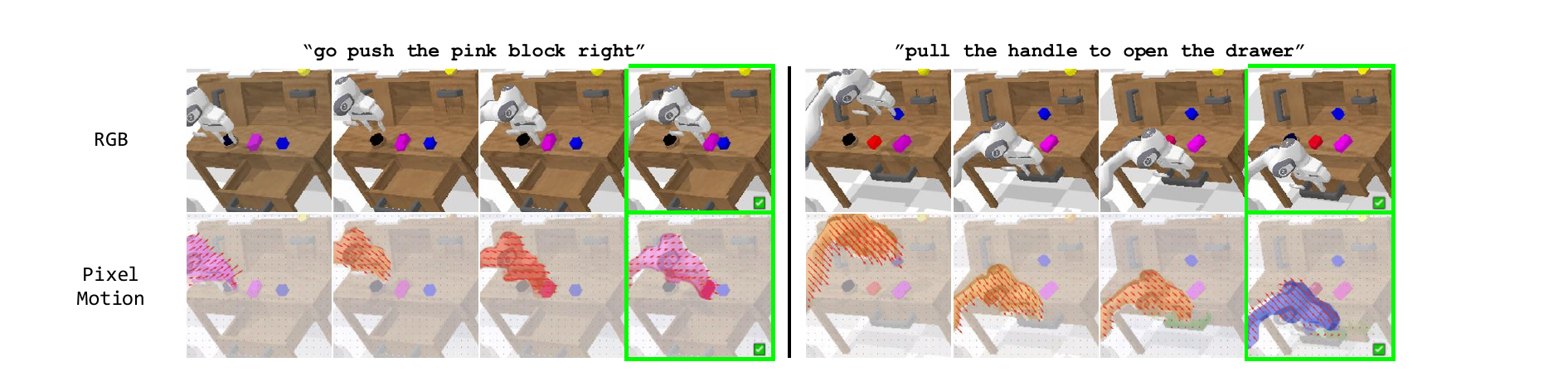}
    \vspace{-1.5em}
    \caption{\textbf{CALVIN rollout examples}. Two example rollouts of \modelname in CALVIN environment. The first row is the sequence of RGB images, and the second row is the visualization of the corresponding pixel motions predicted by \systwoname.} 
    \label{fig:calvin_fs}
\end{figure*}

\subsection{CALVIN Rollout}
We also include additional CALVIN rollout examples with paired RGB frames and predicted pixel motions. Similar to the real-world experiments, \systwoname produces clean, directional pixel-motion fields, and \sysonename executes them through temporally coherent low-level actions. These long-horizon sequences further confirm the consistency between planned and executed motion, even when the tasks involve multi-object interactions, distractors, or ambiguous scene layouts. Our two example rollouts are presented in \Cref{fig:calvin_fs}.

\end{document}